\documentclass[a4paper,conference]{IEEEtran}
\ifCLASSINFOpdf
   \usepackage[pdftex]{graphicx}
\else
\fi
\usepackage{amsmath}
\usepackage{algorithmic}

\usepackage{array}

\ifCLASSOPTIONcompsoc
 \usepackage[caption=false,font=normalsize,labelfont=sf,textfont=sf]{subfig}
\else
 \usepackage[caption=false,font=footnotesize]{subfig}
\fi

\usepackage{xspace}
\usepackage{booktabs}
\usepackage{bm}
\usepackage{amssymb}
\usepackage{multirow}
\usepackage{xcolor}
\usepackage{marvosym}
\makeatletter
\DeclareRobustCommand\onedot{\futurelet\@let@token\@onedot}
\def\@onedot{\ifx\@let@token.\else.\null\fi\xspace}

\def\eg{\emph{e.g}\onedot} 
\def\ie{\emph{i.e}\onedot}

\def\etal{\emph{et al}\onedot}
\makeatother

\begin{document}
%
\title{IterVM: Iterative Vision Modeling Module for\\ Scene Text Recognition}

\author{\IEEEauthorblockN{Xiaojie Chu}
\IEEEauthorblockA{Wangxuan Institute of\\ Computer Technology\\ Peking University, Beijing, China\\
Email: chuxiaojie@stu.pku.edu.cn}
\and
\IEEEauthorblockN{Yongtao Wang\Letter}
\IEEEauthorblockA{Wangxuan Institute of\\ Computer Technology\\ Peking University, Beijing, China\\
Email: wyt@pku.edu.cn}
}


%


\maketitle

\begin{abstract}
Scene text recognition (STR) is a challenging problem due to the imperfect imagery conditions in natural images. State-of-the-art methods utilize both visual cues and linguistic knowledge to tackle this challenging problem. Specifically, they propose iterative language modeling module (IterLM) to repeatedly refine the output sequence from the visual modeling module (VM). Though achieving promising results, the vision modeling module has become the performance bottleneck of these methods. 
In this paper, we newly propose iterative vision modeling module (IterVM) to further improve the STR accuracy. 
Specifically, the first VM directly extracts multi-level features from the input image, and the following VMs re-extract multi-level features from the input image and fuse them with the high-level (i.e., the most semantic one) feature extracted by the previous VM. By combining the proposed IterVM with iterative language modeling module, we further propose a powerful scene text recognizer called IterNet. Extensive experiments demonstrate that the proposed IterVM can significantly improve the scene text recognition accuracy, especially on low-quality scene text images. Moreover, the proposed scene text recognizer IterNet achieves new state-of-the-art results on several public benchmarks. Codes will be available at https://github.com/VDIGPKU/IterNet.

\end{abstract}


%
\IEEEpeerreviewmaketitle

\section{Introduction}
Reading text in scene images is the basis of various applications, such as multilingual translation, blind navigation, and automatic driving. 
As a result, scene text recognition (STR) has attracted a lot of research interests in recent years. Scene text recognition is more difficult than document text recognition because of the imperfect imagery conditions in natural images, such as complex backgrounds, low resolution, perspective, curved distortion. 
To tackle these challenges, recent deep learning based works utilize visual modeling module and language modeling module for extracting both visual and linguistic information contained in the scene text image.

\begin{figure}[t]
\centering
    \centering
    \subfloat[Vision modeling module (VM) with decoder~\cite{shi2018aster}.]{
    \includegraphics[width=0.95\linewidth]{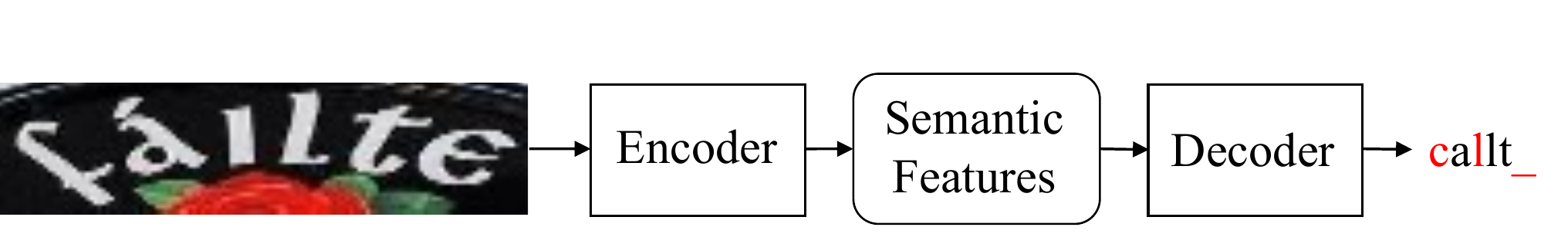}
    }\label{fig:exp_VM}\\
    \vspace{-1em}
    \subfloat[VM with iterative language modeling module (IterLM)~\cite{yu2020towards,fang2021read}.]{
    \includegraphics[width=0.95\linewidth]{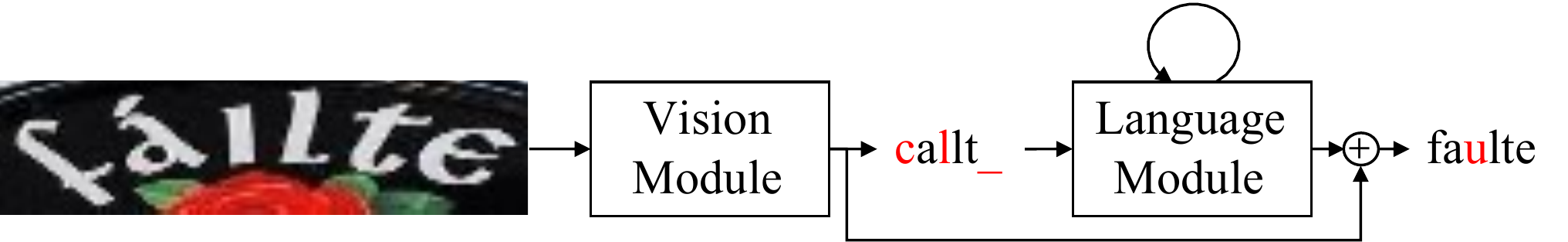}
    }\label{fig:exp_IterLM}\\
    \vspace{-1em}
    \subfloat[IterVM (ours): Iterative vision modeling module.]{
    \includegraphics[width=0.95\linewidth]{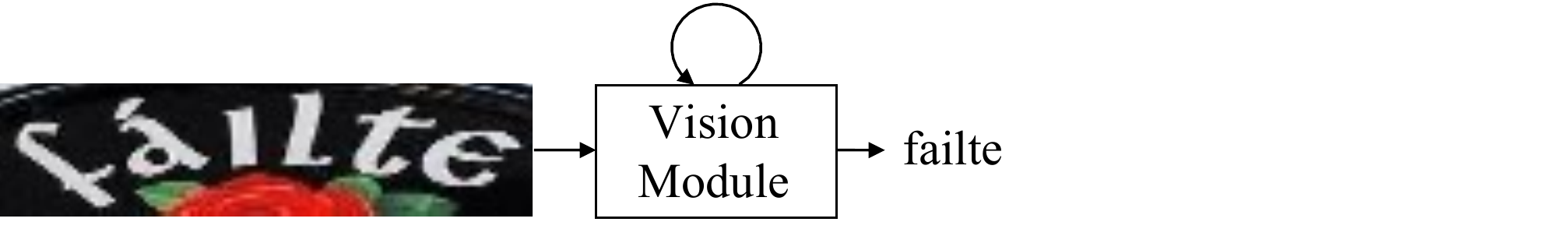}
    }\label{fig:exp_IterVM}\\
    \vspace{-1em}
    \subfloat[IterNet (ours): IterVM with IterLM.]{
    \includegraphics[width=0.95\linewidth]{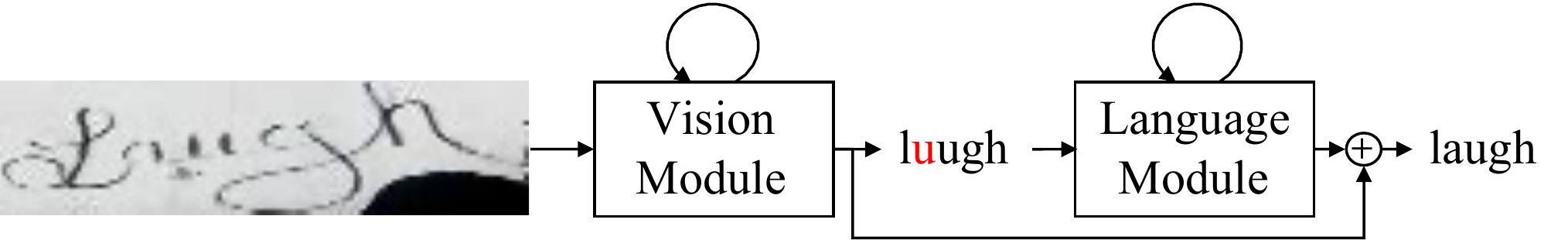}
    }\label{fig:exp_IterNet}
\caption{Comparison between different modules in scene text recognizer.}
\vspace{-.0em}
\label{fig:example}
\end{figure}

The vision modeling module (VM) typically consists of an encoder that extracts high-level semantic information from input images and a decoder for character sequence generation, as illustrated in Fig.~\ref{fig:example}(a) and Fig.~\ref{fig:IterVM}. Many recent works attempt to improve the performance of scene text recognition from the perspective of extracting more robust and effective visual features, such as upgrading the backbone networks~\cite{wang2017gated, cheng2017focusing, li2019show}, adding rectification module for irregular text images \cite{shi2016robust,shi2018aster,yang2019symmetry} and improving attention mechanisms~\cite{yang2017learning, wang2020decoupled}. Though achieving promising results, these methods are designed for specific scenarios (\eg, irregular or low-resolution scene text images) or focus on high-level semantic reasoning between visual clues and text characters. In other words, there is still room for improvement in their visual feature extraction process from low-level to high-level.

In order to take advantage of linguistic knowledge, newly proposed STR methods~\cite{yu2020towards, fang2021read} exploit an iterative language modeling module (IterLM) that repeatedly refines the output sequence of VM, and combine the outputs of LM and VM to produce the final prediction. Under such a pipeline, the vision modeling module becomes the performance bottleneck. For example, as illustrated in Fig.~\ref{fig:example}(b), it is difficult for the IterLM to correct the bad prediction of the VM.

In this paper, we focus on the critical process of extracting visual features for scene text recognition.
Inspired by the human behavior of repeatedly viewing and comparing when recognizing difficult text characters, we propose an iterative approach that allows vision modeling module to repeatedly extract and enhance visual features from the input scene text image. In particular, as shown in Fig.~\ref{fig:example}(c),
we introduce feedback connections from the output into the middle layers of vision modeling module.
In this way, we make the vision modeling module look at the input scene text image twice or more times by unrolling the recursive structure to an iterative implementation.
Specifically, at the first iteration, the VM extracts high-level semantic feature directly from the input image. While at the subsequent iterations, the VM extracts and refines features with the help of high-level feature obtained from the previous iteration. 
Similar to the iterative language modeling module (IterLM) in ABINet~\cite{fang2021read}, our IterVM can iteratively enhance the visual features for scene text recognition.

Furthermore, the proposed IterVM works very well with the iterative language modeling module (IterLM)~\cite{fang2021read}. To be more specific, we propose a powerful scene text recognizer named IterNet, by combing both IterVM and IterLM. As illustrated in Fig.~\ref{fig:example}(d), IterNet iteratively extracts and enhances visual features with IterVM, and then repeatedly updates the results with linguistic information using IterLM. Experimental results show that IterNet inherits the benefits of both IterVM and IterLM, thus can handle more challenging cases. Taking the hard case shown in Fig.~\ref{fig:example}(d) as an example, the IterVM correctly recognizes all the characters except for the second one (\ie, the second character 'a' is wrongly recognized as 'u'), and then IterLM further corrects this error with the linguistic knowledge that it could be inferred as the word 'laugh'.

To sum up, the main contributions of this paper are as follows:
\begin{itemize}
\item 
We propose IterVM, an iterative approach for visual feature extraction which can significantly improve scene text recognition accuracy.

\item By combining the proposed IterVM with IterLM, we propose a powerful scene text recognizer named IterNet and achieve new state-of-the-art results on several public benchmarks.
\end{itemize}

\section{Related Work}
\subsection{Overview of Scene Text Recognition (STR)}
Recent STR methods mainly exploit the encoder-decoder framework for mapping the entire text image into a target string sequence~\cite{chen2021text}.
CRNN~\cite{shi2017end} uses VGG~\cite{Simonyan15} and bidirectional LSTM (BiLSTM) to extract visual features from the image and adopt the CTC-based decoder to predict character sequence. 
Attention-based decoders, which learn the alignment between the encoded visual feature vectors and the target characters, are introduced by~\cite{shi2016robust} and~\cite{lee2016recursive} to improve the recognition accuracy.
Our method also follows the attention-based encoder-decoder architecture.

\subsection{Visual Feature Extraction for STR}
A large category of existing works make efforts to improve the recognition accuracy by mitigating the interferences from imperfect imaging conditions with some sophisticatedly designed image preprocessing modules, \eg, rectification module~\cite{shi2016robust, liu2016star, liu2018char, yang2019symmetry, luo2019moran, zhan2019esir} for geometric distortion and chromatic distortion~\cite{DBLP:conf/aaai/0003XCPNWZ21}, super-resolution unit~\cite{yanplugnet} for low resolution.
Another category of existing methods enhance spatial visual features to handle irregular text images. AON~\cite{cheng2018aon} proposes an arbitrary orientation network to capture character features in four directions. SAR~\cite{li2019show} proposes a 2D tailored attention module to tackle the challenge of irregular scene text recognition. SATRN~\cite{lee2020recognizing} uses 2D self-attention mechanism to capture the spatial dependency in 2D feature map for recognizing texts with arbitrary shapes. RobustScanner~\cite{yue2020robustscanner} introduces a position enhancement branch and a dynamic fusion module to mitigate the errors in contextless scenarios. To address the misalignment issue, DAN~\cite{wang2020decoupled} decouples the decoder of the traditional attention mechanism into a convolutional alignment module and a decoupled text decoder.
Moreover, for more representative visual features, some STR methods propose complex backbone networks (\eg, variant of ResNet~\cite{liu2016star, shi2018aster, yu2020towards}). More recently, AutoSTR~\cite{zhang2020efficient} uses neural architecture search technique to find a data-dependent backbone (\ie, visual feature extractor) for STR. 
Unlike the above mentioned methods, which design novel modules or backbone networks, our IterVM novelly exploits an iterative approach to enhance the visual features.

\subsection{Visual Feature Enhancement by Semantic Information}
To further improve the recognition accuracy, existing STR methods usually use semantic information to enhance the visual feature by introducing auxiliary supervision or designing new modules.
EFIFSTR~\cite{wang2020exploring} uses attentional glyph generation with trainable font embeddings for learning font-independent features of scene texts.
SCATTER~\cite{litman2020scatter} adds intermediate supervisions to train a deeper BiLSTM encoder. 
SEED~\cite{qiao2020seed}~invents semantics enhanced encoder-decoder framework which predicts an additional global semantic information supervised by the word embedding from a pre-trained language model.
VisionLAN~\cite{wang2021two} proposes language-aware visual masks for training, which simulates the case of missing character-wise visual semantics and guides the vision modeling module to use not only the visual texture of characters but also the linguistic information in visual context for recognition.
PREN2D~\cite{yan2021primitive} proposes global feature aggregations to learn primitive visual representations from multi-scale feature maps and exploits GCNs to transform primitive representations into high-level visual text representations.
Different from these works, our IterVM uses feedback connections to fuse high-level (the most semantic) visual feature with multi-level visual features.

\subsection{Explicit Language Modeling for STR}
Existing methods~\cite{fang2018attention, yu2020towards, fang2021read} usually make use of linguistic knowledge to boost the recognition accuracy by refining the output sequence of the VM with it. Fang~\etal~\cite{fang2018attention} use gated convolutional layers to model the language sequences at the character level. SRN~\cite{yu2020towards} proposes a global semantic reasoning module based on transformer units~\cite{vaswani2017attention} for pure language modeling. ABINet~\cite{fang2021read} proposes an iterative correction scheme with the language model, which progressively refines the recognition results. Bhunia~\etal~\cite{bhunia2021joint} propose a multi-stage multi-scale decoder that refers to visual features multiple times to enhance linguistic features.  PIMNet~\cite{qiao2021pimnet} adopts an iterative easy-first~\cite{goldberg2010efficient} decoding strategy, which predicts the most confident characters first in each iteration and predicts the remaining ones in the next iteration. In contrast to these works that has only explored the iterative principle for language modeling, we newly perform the iterative method for both language and vision modeling. Experiment results show that our IterVM can work well together with the iterative language modeling module.

\section{Method}
In this section, we first detail the vision modeling module used in this work in Sec.~\ref{sec:VM}, and then introduce our IterVM and IterNet in Sec.~\ref{sec:IterVM} and Sec.~\ref{sec:IterNet} respectively. We will finally introduce the supervision used for training in Sec.~\ref{sec:loss}.

\subsection{Vision Modeling Module (VM)} \label{sec:VM}
Since the vision modeling module (VM) is designed to recognize texts as a stand-alone model, we follow the encoder-decoder framework for STR task, as illustrated in Fig.~\ref{fig:example}(a).
The VM is used as encoder, which encodes input images and reasons semantic information, and a decoder is employed for converting high-level visual features into sequence of characters.
Following ABINet~\cite{fang2021read}, we use ResNet and Transformer units as encoder and a position attention based prediction layer as decoder.

In detail, the ResNet~\cite{shi2018aster, wang2020decoupled, fang2021read} has $5$ residual blocks in total, with down-sampling after the first and third blocks. 
In detail, a convolution with $3 \times 3$ kernel size encodes an input image $\bm{x}$ of size $H \times W \times 3$ into a feature map $\mathbf{X}_0$. Then, we have:
\begin{equation}
\mathbf{X}_k = \mathcal{R}_k(\mathbf{X}_{l-1}) \in \mathbb{R}^{H_k \times W_k \times C_k},
\label{eq:resnet}
\end{equation}
where $k=1,2,...,5$ and $H_k,W_k$ are the spatial size of $\mathbf{X}_k$ and $C_k$ is feature dimension. The output of ResNet is $\mathbf{X}_n \in \mathbb{R}^{\frac{H}{4} \times \frac{W}{4} \times C}$. Then, Transformer units~\cite{yu2020towards, fang2021read} are employed for sequence modeling:
\begin{equation}
\mathbf{S} = \mathcal{T}(\mathbf{X}_n) \in \mathbb{R}^{\frac{HW}{16} \times C}.
\label{eq:transformer}
\end{equation}
Finally, an position attention based decoder~\cite{fang2021read} transcribes visual features into character probabilities:
\begin{equation}
\mathbf{Y}_v = \mathcal{D}(\textbf{S}) = \text{softmax}(\frac{\mathbf{Q}\mathcal{G}(\mathbf{S})^\mathsf{T}}{\sqrt{C}})\mathbf{S}\mathbf{W},
\label{eq:decoder}
\end{equation}
where $\mathbf{Q} \in \mathbb{R}^{T \times C}$ is positional encodings~\cite{vaswani2017attention} of character orders ($T$ is the length of character sequence), $\mathcal{G}(\cdot)$ is a mini U-Net, $\mathcal{G}(\mathbf{S}) \in \mathbb{R}^{\frac{HW}{16} \times C}$, and
$\mathbf{W} \in \mathbb{R}^{C \times D}$ denotes a linear transition matrix ($D$ represents the number of character classes).

\subsection{Iterative Vision Modeling Module (IterVM)} \label{sec:IterVM}

\begin{figure}[t]
    \centering
    \includegraphics[width=1\linewidth]{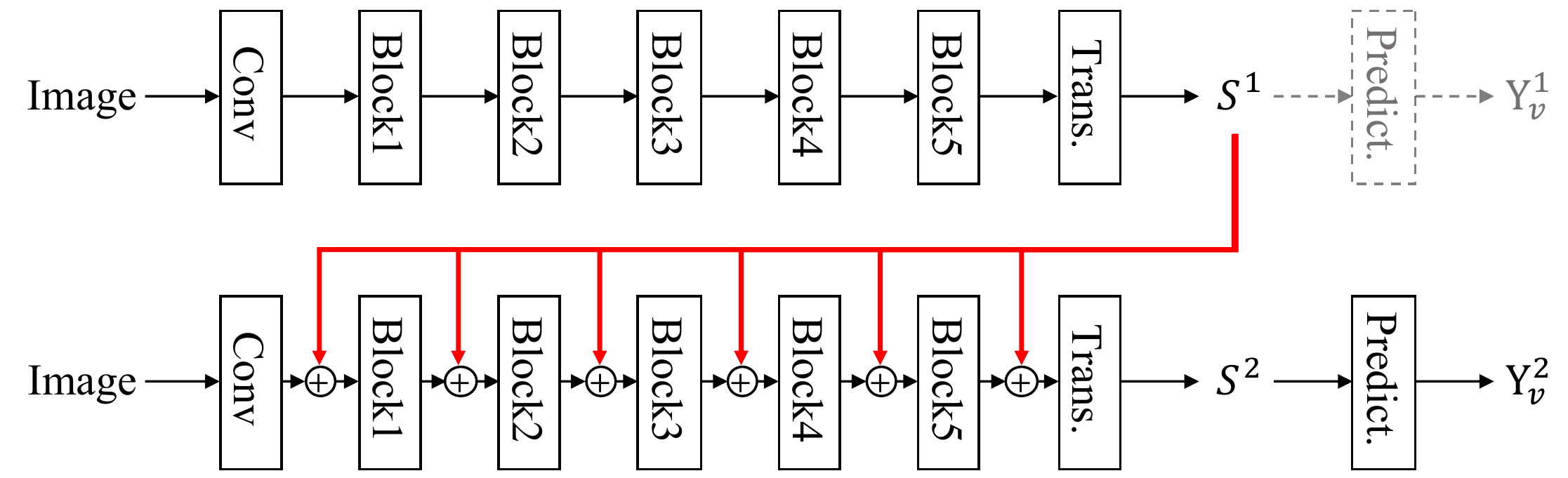}
\caption{The architecture of iterative vision modeling module (IterVM) with 2 iterations. The red line indicates the feedback connections, and the dotted line indicates the process which is only executed for training. \textit{Trans.} represents operations in Transformer and  \textit{Predict.} represents the prediction layer which is shared across all iterations.}
\label{fig:IterVM}
\vspace{-.0em}
\end{figure}

Like all STR models, our proposed iterative vision modeling module (IterVM) has an encoder that maps the input image to semantic feature and a decoder that predicts texts based on the extracted semantic feature. Unlike classical vision modeling modules, we adopt an iterative design that allows the encoder to extract visual features and refine them iteratively. 

Specifically, our IterVM adds feedback connections to the encoder as illustrated in Fig.~\ref{fig:example}(c) and Fig.~\ref{fig:IterVM}. 
In other words, the encoder of VM is executed $N$ times with the same input image but different semantic feature for fusion. 
Let $\mathcal{R}^i$ and $\mathcal{T}^i$ denote the operations of ResNet and Transformer at iteration $i$, respectively.
For the first iteration, on the basis of Eq.~\ref{eq:resnet} and Eq.~\ref{eq:transformer}, $\mathcal{R}^1$ and $\mathcal{T}^1$ extract visual features from input images directly and the output of high-level semantic feature is marked as $\mathbf{S}^1$. 
When at the iteration $i \geq 2$, the original inputs of $\mathcal{R}^i$ and $\mathcal{T}^i$ are supplemented with semantic feature as follows:
\begin{equation}
\mathbf{X}^i_k =\mathcal{R}^i_k(\mathbf{X}^i_{k-1}+\mathcal{U}(\mathbf{W}^i_{k-1}\mathbf{S}^{i-1})),
\end{equation}
\begin{equation}
\mathbf{S}^i = \mathcal{T}^i(\mathbf{X}^i_n+\mathbf{W}^i_{n}\mathbf{S}^{i-1})),
\end{equation}
where $\mathbf{W}^i_{k-1} \in \mathbb{R}^{C \times C_{k-1}}$ indicates a linear transition matrix for aligning hidden dimension of features and $\mathcal{U}$ represents upsample operation for aligning the spatial dimensions. 

During training, on the basis of Eq.~\ref{eq:decoder}, all visual semantic features $\mathbf{S}^i$ are transcribed into character sequences by the same decoder as follows:
\begin{align}
\mathbf{Y}^i_v = \mathcal{D}(\mathbf{S}^i).
\end{align}
All predictions $\mathbf{Y}^i_v$ are used to calculate the loss for supervision. 
During testing, only the prediction $\mathbf{Y}^N_v$ at the last iteration (\ie, $i=N$) is taken as the final output of IterVM, and the decoding process of the semantic feature at the intermediate iterations can be skipped to save inference time.

\subsection{IterNet} \label{sec:IterNet}

\begin{figure}[t]
    \centering
    \includegraphics[width=1\linewidth]{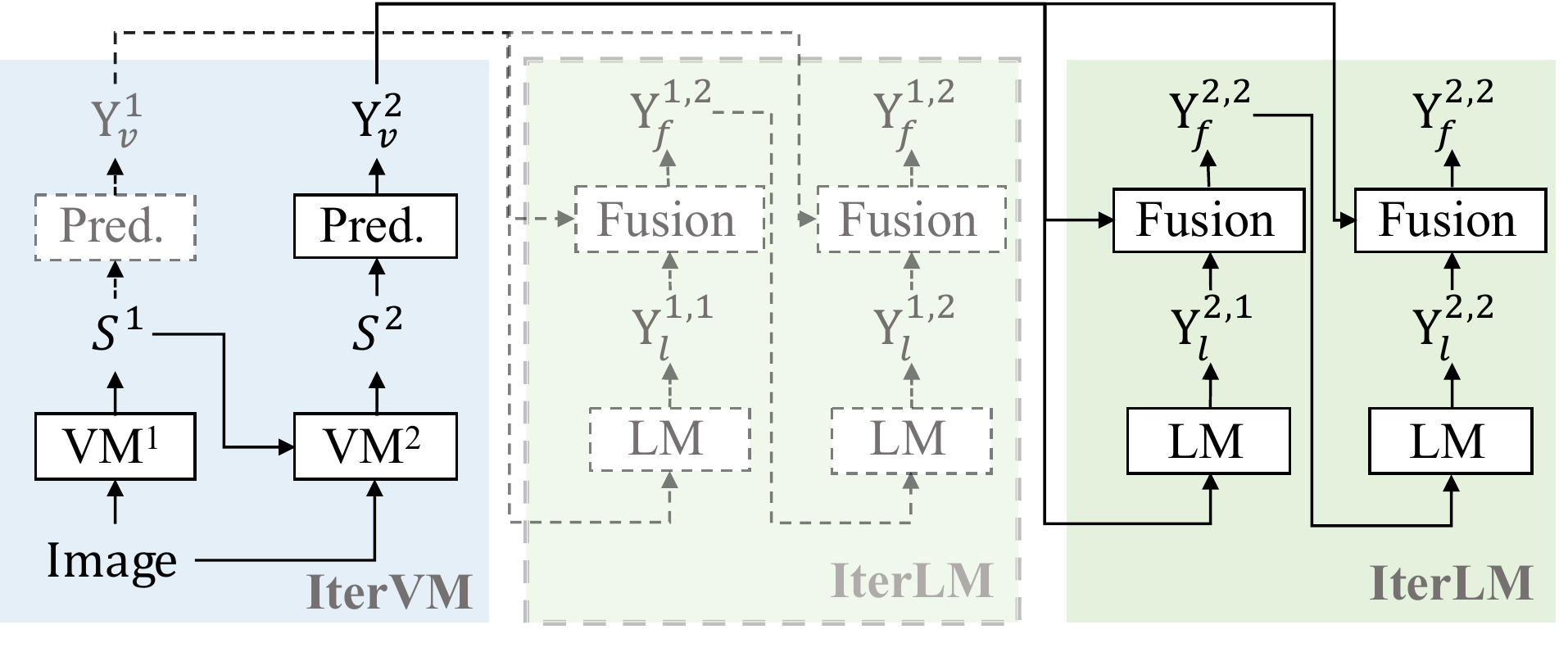}
\caption{The architecture of our IterNet, which consists of iterative vision modeling module (IterVM) and iterative language modeling module (IterLM). We unroll IterNet to an example of sequential implementation with 2-iteration VM and 2-iteration LM. 
The dotted line indicates the process which is only executed for training. \textit{Pred.} represents the prediction layer which is shared across all iterations. }
\vspace{-.0em}
\label{fig:IterNet}
\end{figure}

To use linguistic knowledge as a complement to visual information, we use iterative language modeling module (IterLM) to further correct the prediction results of IterVM and obtain a powerful recognizer called IterNet.
The architecture of the IterNet is shown in Fig.~\ref{fig:IterNet}. 

Firstly, IterVM iteratively extracts visual features $S^i$ from the input image and outputs text predictions $Y^i$ solely based on visual information.
Then, the IterLM execute $M$ times with different input:
\begin{equation}
\mathbf{Y}^{i,j}_l = \mathcal{LM}(\mathbf{Y}_f^{i,j-1}),
\end{equation}
where $i=1,2,..,N$, $j=1,2,...,M$ and $\mathcal{LM}$ indicates the share-weight IterLM.
For the first iteration (\ie, $j=1$), the input of IterLM is the prediction $\mathbf{Y}_v$ from our IterVM (\ie, $\mathbf{Y}_f^{i,0} = \mathbf{Y}^i_v$). For the following iterations, the input is the fusion of $\mathbf{Y}^i_v$ and $\mathbf{Y}^{i,j}_l$ from the previous iteration:
\begin{equation}
    \mathbf{Y}_f^{i,j} = \mathcal{F}(\mathbf{Y}^{i,j}_l, \mathbf{Y}^i_v),
\end{equation}
where $\mathcal{F}$ is gating operation~\cite{yu2020towards, yue2020robustscanner, fang2021read} for the final decision.
During the training stage, IterLM corrects all the IterVM predictions $\mathbf{Y}^i_v$ and calculates losses for supervision. 
In the testing stage, only the output $\mathbf{Y}^N_v$ of IterVM at the last iteration (\ie, $i=N$) is used as the input for IterLM and the output $\mathbf{Y}_f^{N,M}$ of IterLM at the last iteration (\ie, $j=M$) is taken as the final output of IterNet.

\subsection{Supervised Training} \label{sec:loss}
To improve the training of our IterVM, we propose to add intermediate supervision to all semantic features extracted by vision modeling module at all iterations.
Specifically, all semantic features are decoded to predictions by a weight sharing decoder, and all these predictions are refined by a weight sharing IterLM for calculating loss.

IterVM is trained end-to-end using the following multi-task objectives:
\begin{equation}
\mathcal{L}_v = \sum^N_{i=1}{\mathcal{L}^i_v},
\end{equation}
where $\mathcal{L}^i_{l}$ denotes the cross-entropy losses from $\mathbf{Y}^{i}_{v}$ at $i$-th iteration. While IterNet is trained end-to-end with the following multi-task objectives:
\begin{equation}
\mathcal{L} = \sum^N_{i=1}{\mathcal{L}^i_v} + \sum^N_{i=1}{\sum^M_{j=1}{(\mathcal{L}^{i,j}_l + \mathcal{L}^{i,j}_f)}},
\label{eq:loss}
\end{equation}
where $\mathcal{L}^{i,j}_{l}$ and $\mathcal{L}^{i,j}_{f}$ are the cross-entropy losses from $\mathbf{Y}^{i,j}_{l}$ and $\mathbf{Y}^{i,j}_{f}$, respectively.

\section{Experiments}
\subsection{Datasets and Evaluation} \label{sec:datasets}
For fair comparisons, we use the same training datasets and evaluation protocol as~\cite{qiao2020seed, yu2020towards, fang2021read, wang2021two}. In detail,  we use two widely used synthetic datasets for the training: MJSynth~\cite{jaderberg2014synthetic} and SynthText~\cite{gupta2016synthetic}. 
MJSynth contains 9 million synthetic text images, while SynthText contains 7 million images, including examples with special characters. Most previous works have used these two synthetic datasets together for training~\cite{baek2019wrong}.

For evaluation, six widely used real-world STR test datasets are used as benchmarks. 
These datasets are classified as either regular or irregular based on the difficulty and geometric layout of the texts~\cite{baek2019wrong}.

\textbf{Regular} datasets mainly contain horizontally aligned text images. IIIT5K (IIIT)~\cite{mishra2012scene} is made up of 3,000 images gathered from the Internet. Street View Text (SVT)~\cite{wang2011end} contains 647 images from Google Street View. ICDAR2013 (IC13)~\cite{karatzas2013icdar} was created for the ICDAR 2015 Robust Reading competitions and it has two variants: 857 images (IC13$_{S}$) and 1,015 images (IC13$_{L}$). We use both variants for fair comparisons.

\textbf{Irregular} datasets consist of more examples of text in arbitrary shapes. 
ICDAR2015 (IC15)~\cite{karatzas2015icdar} contains images taken from daily scenes and also has two versions: 1,811 images (IC15$_{S}$) and 2,077 images (IC15$_{L}$). Street View Text Perspective (SVTP)~\cite{quy2013recognizing} includes 645 images with text captured in perspective views. CUTE80~(CUTE)~\cite{risnumawan2014robust} consists of 288 images with heavily curved text. Details about these datasets can be found in previous works~\cite{qiao2020seed, yu2020towards}. 

In all of our experiments, recognition performance is measured by case-insensitive alphanumeric accuracy in the lexicon-free mode~\cite{shi2018aster,baek2019wrong}.
We also evaluate the accuracy of overall datasets except for IC13$_{L}$ and IC15$_{L}$ and report it for comparison. 

\subsection{Implementation Details}\label{sec:Implementation_detail} 

\subsubsection{Architecture Details}
We follow the implementation of architecture from ABINet~\cite{fang2021read}. 
The model size is set to 512. To reduce computational costs, we use ResNet-25 instead of ResNet-45. In particular, we reduce the number of residual units in each block in the original ResNet-45~\cite{fang2021read} from [3, 4, 6, 6, 3] to [2, 2, 3, 3, 2]. We employ the iterative language modeling module (IterLM) in ABINet~\cite{fang2021read}, and it consists of 3 Transformer decoder blocks. We initialize the IterLM with the pre-trained weights provided by~\cite{fang2021read}.
Unless otherwise specified, the iteration number $N$ for IterVM and $M$ for IterLM are set to 3.

\subsubsection{Training Details}
Images are directly resized to $32 \times 128$ with data augmentation used in~\cite{fang2021read}. All models are trained end-to-end with a total batch size of 384.
ADAM optimizer is adopted with the initial learning rate $1e^{-4}$, which is decayed to $1e^{-5}$ after 6 epochs.  To save time in training, we only use 20\% of the training data in the ablation study.

\subsection{Ablation Study}
We perform a series of ablation experiments to demonstrate the effectiveness of the proposed IterVM, IterNet, and intermediate supervision.
\subsubsection{Effectiveness of IterVM}
\begin{table}[]
\centering
\caption{Ablation study of the number of iterations for IterVM.}
\vspace{-.0em}
\label{tab:iterative}
\begin{tabular}{c|cc|cc|c}
\toprule
\multirow{2}{*}{Model}  & \multicolumn{2}{c|}{\#   Iteration} & \multicolumn{3}{c}{Accuracy}                                   \\
                           &VM  & LM & Regular  & Irregular   & Total         \\
\midrule
\multirow{4}{*}{IterVM} & 1          &-                    & 90.3       & 74.1       & 84.2  \\
\cmidrule{2-6}
&2       &-                       & 92.5$^{+2.2}$ & 80.1$^{+6.0}$ & 87.8$^{+3.6}$ \\
&3        &-                      & 93.8$^{+3.5}$ & 82.7$^{+8.6}$ & 89.6$^{+5.4}$ \\
&4          &-                    & 93.9$^{+3.6}$ & 83.8$^{+9.7}$ & 90.1$^{+5.9}$ \\
\midrule
\multirow{3}{*}{IterNet} &1 & 3 & 94.5& 83.6& 90.4         \\
\cmidrule{2-6}
&2 & 3& 95.4$^{+0.9}$            & 86.0$^{+2.4}$            & 91.9$^{+1.5}$ \\
&3 &3& 96.1$^{+1.6}$            & 86.8$^{+3.2}$            & 92.6$^{+2.2}$ \\
\bottomrule
\end{tabular}
\end{table}

We investigate the impact of different iterations (from 1 to 4) of IterVM on the model performance. 
It is worth noting that \textit{the vision modeling module with only 1 iteration is the baseline}, that is, iteration operation for VM is not performed.

As demonstrated by the results in Table~\ref{tab:iterative}, our IterVM offers significant performance improvements compared to the baseline.
The more iterations for IterVM, the better performance.
Compared to the baseline, the IterVM with two iterations can improve the accuracy on regular and irregular datasets by $2.2\%$ and $6.0\%$, respectively. When the number of iterations increases further, the performance on regular and irregular datasets can be improved by $3.5\%\sim3.6\%$ and $8.6\%\sim9.7\%$, respectively, resulting in an overall improvement of $5.4\%\sim5.9\%$.
These results indicate that the iterative approach can effectively improve the capability of vision modeling.

\subsubsection{Compatibility of IterVM with IterLM}
We further conduct experiments on IterNet to verify the effect of the iterative manner for vision modeling module on the final recognition accuracy when the iterative language modeling module is used.
It’s worth noting that \textit{the baseline in this experiment is the ABINet~\cite{fang2021read}, in which the number of iterations for VM is 1}. As shown in Table~\ref{tab:iterative}, our IterVM still achieves significant gains (\ie, $2.4\%\sim3.2\%$ on irregular datasets) over the baseline. These results demonstrate that the proposed IterVM can work well together with IterLM.

\subsubsection{Comparison on computational resources}
\begin{figure}[t]
    \centering
    \includegraphics[width=1\linewidth]{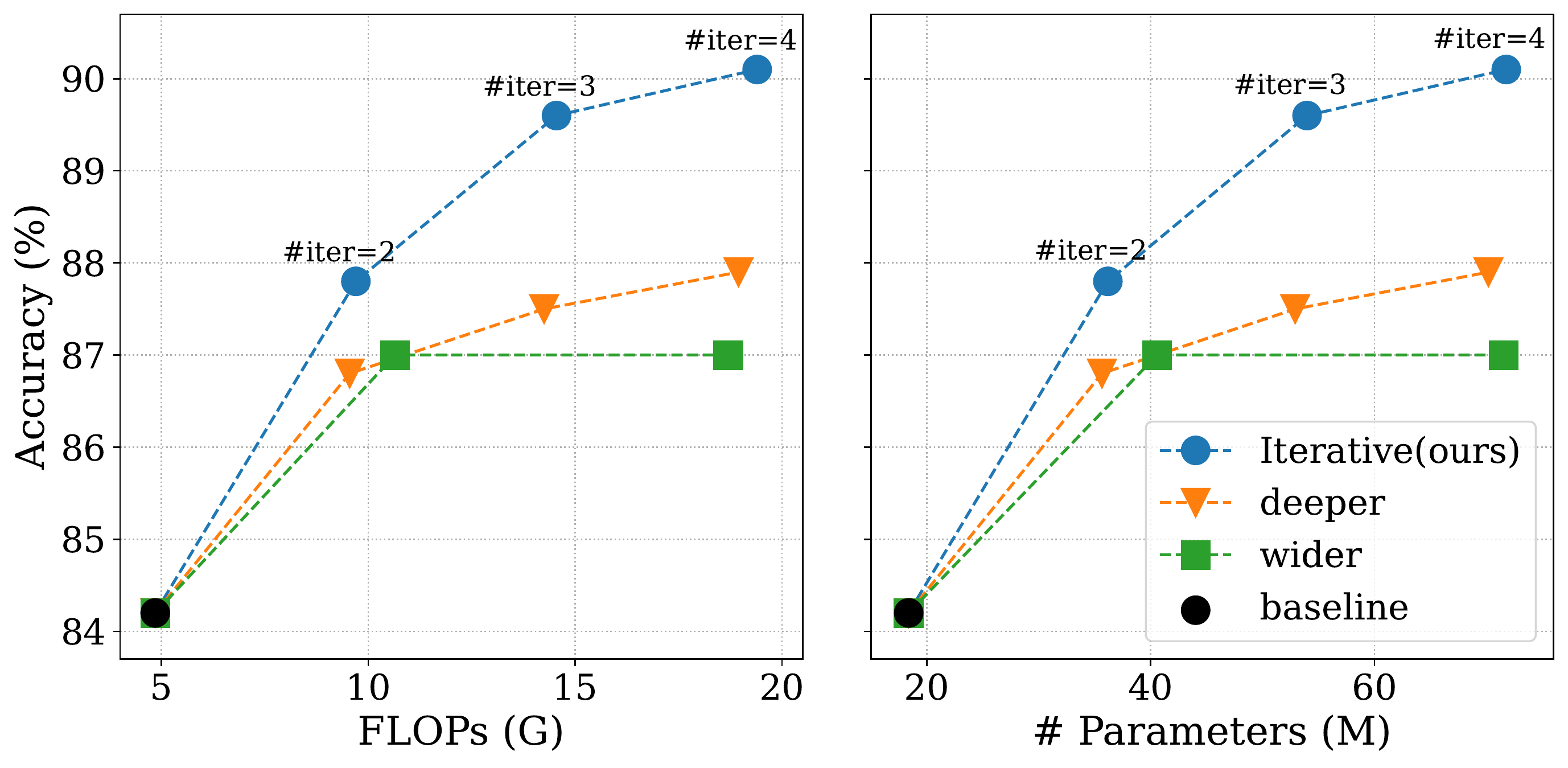}
\caption{Comparison on computation cost (FLOPs) and model size (\#Parameters). \textbf{Left}: Computation cost vs. Accuracy; \textbf{Right}: Model size vs. Accuracy. Our IterVM significantly outperform deeper or wider ones. }
\vspace{-.0em}
\label{fig:pareto}
\end{figure}

It is common knowledge that more computations and parameters result in better accuracy.
Since the computation cost of IterVM is scaled linearly with the number of iterations, we compare the IterVM with deeper and wider baselines to ensure a fair comparison.  For deeper baselines, the depth of vision modeling module is $2\times$, $3\times$, and $4\times$, respectively. For wider baselines, the channels of the vision modeling module are $1.5 \times$ and $2 \times$, respectively. For IterVM, the number of iterations ranges from 2 to 4.

As illustrated in Fig~\ref{fig:pareto}, IterVM with any iteration is more accurate than the deeper and wider baselines. As shown in the left of Fig~\ref{fig:pareto}, when FLOPs is larger than 14G, IterVM shows huge performance gains of more than 2\% under the same computation cost. 
Compared to a $4\times$ deeper baseline, our IterVM with 2 iterations achieves comparable performance with approximately $2\times$ fewer FLOPs. 
The same phenomenon can be observed when comparing the number of parameters (in the right of Fig~\ref{fig:pareto}): our IterVM achieves higher accuracy with fewer parameters. 
These results indicate that: our proposed iterative manner for the vision modeling module is more effective and efficient than simply making the network deeper or wider.

\subsubsection{Effectiveness of intermediate supervision}
\begin{table}
\vspace{-.0em}
\caption{
Ablation study of the intermediate supervision.
We report the total accuracy of the predictions by IterVM ($\mathbf{Y}_v$), IterLM ($\mathbf{Y}_l$), and IterNet($\mathbf{Y}_f$).
}
\label{tab:loss}
\centering
\begin{tabular}{c|cc|c}
\toprule
Intermediate & \multicolumn{3}{c}{Accuracy} \\
Supervision & IterVM ($\mathbf{Y}_v$) & IterLM ($\mathbf{Y}_l$) & IterNet ($\mathbf{Y}_f$) \\
\midrule
 & 89.0 & 49.7 & 92.1 \\
${\surd}$ & 89.9$^{+0.9}$ & 50.2$^{+0.5}$ & 92.6$^{+0.5}$  \\
\bottomrule
\end{tabular}
\vspace{-.0em}
\end{table}
As introduced in Sec.~\ref{sec:loss}, we introduce intermediate supervision to improve the training of our IterVM, thus we conduct experiments on IterNet to verify its effectiveness. 
As shown in Table~\ref{tab:loss}, with the help of intermediate supervision, the results of $\mathbf{Y}_v$ (output from IterVM), $\mathbf{Y}_l$ (output from IterLM) and $\mathbf{Y}_f$ (output from the IterNet) are improved by 0.9\%, 0.5\% and 0.5\% respectively. 
These results show that the proposed intermediate supervision can effectively boost the performance of IterNet as well as the modules in it, especially for IterVM.

\subsection{Comparison to State-of-the-Arts}
\begin{table*}
\caption{Recognition accuracy comparisons with state-of-the-art methods on six benchmark datasets (including variant versions of IC13 and IC15). 
The best performing results are shown in bold font, and the second-best results are shown in the underlined font.
}
\vspace{-0.5em}
\label{tab:accuracy}
\centering
\begin{tabular}{ll|cccc|cccc}
\toprule
 & & \multicolumn{4}{c|}{\textbf{Regular test dataset}} & \multicolumn{4}{c}{\textbf{Irregular test dataset}} \\
 \textbf{Model} & \textbf{Year} & IIIT & SVT & IC13$_{S}$ & IC13$_{L}$ & IC15$_{S}$ & IC15$_{L}$ & SVTP & CUTE \\
\midrule
 TRBA~\cite{baek2019wrong} & 2019 & 87.9 & 87.5 & 93.6 & 92.3 & 77.6 & 71.8 & 79.2 & 74.0 \\
 ESIR~\cite{zhan2019esir} & 2019 & 93.3 & 90.2 & - & 91.3 & - & 76.9 & 79.6 & 83.3 \\
 SE-ASTER~\cite{qiao2020seed} & 2020 & 93.8 & 89.6 & - & 92.8 & 80.0 & & 81.4 & 83.6 \\
 DAN~\cite{wang2020decoupled} & 2020 & 94.3 & 89.2 & - & 93.9 & - & 74.5 & 80.0 & 84.4 \\
 RobustScanner~\cite{yue2020robustscanner} & 2020 & 95.3 & 88.1 & - & 94.8 & - & 77.1 & 79.5 & 90.3 \\
 AutoSTR~\cite{zhang2020efficient} & 2020 & 94.7 & 90.9 & - & 94.2 & 81.8 & - & 81.7 & - \\
 SATRN~\cite{lee2020recognizing} & 2020 & 92.8 & 91.3 & - & 94.1 & - & 79.0 & 86.5 & 87.8 \\
 SRN~\cite{yu2020towards} & 2020 & 94.8 & 91.5 & 95.5 & - & 82.7 & - & 85.1 & 87.8 \\
 GA-SPIN~\cite{DBLP:conf/aaai/0003XCPNWZ21} & 2021 & 95.2 & 90.9 & - & 94.8 & 82.8 & 79.5 & 83.2 & 87.5 \\
 PREN2D~\cite{yan2021primitive} & 2021 & 95.6 & \underline{94.0} & 96.4 & - & 83.0 & - & 87.6 & \textbf{91.7} \\
 Bhunia \etal~\cite{bhunia2021joint} & 2021 & 95.2 & 92.2 & - & \underline{95.5} & - & \textbf{84.0} & 85.7 & 89.7 \\
 VisionLAN~\cite{wang2021two} & 2021 & 95.8 & 91.7 & 95.7 & - & 83.7 & - & 86.0 & 88.5 \\
 ABINet~\cite{fang2021read} & 2021 & \underline{96.2} & 93.5 & \underline{97.4} & - & \underline{86.0} & - & \underline{89.3} & 89.2 \\
 \midrule
 IterNet (ours) &  & \textbf{96.8} & \textbf{95.1} & \textbf{97.9} & \textbf{97.0} & \textbf{87.7} & \textbf{84.0} & \textbf{90.9} & \underline{91.3} \\
\bottomrule
\end{tabular}
\end{table*}
We compare the experimental results of our IterNet with the state-of-the-art ones on 6 widely used STR benchmarks (including variant versions of IC13 and IC15). Only the methods trained with MJSynth and SynthText are considered for fair comparisons.

Among the existing methods, PREN2D, Bhunia \etal, and ABINet achieve state-of-the-art results on different benchmarks (see underlined and bold values in the table). When compared to them, our IterNet achieves new state-of-the-art results on all evaluation datasets except CUTE. Specifically, our model achieves superior performance improvements on SVT, IC13$_\text{L}$, IC15$_\text{S}$ and SVTP (datasets contain low-quality images) by 1.1\%$\sim$1.7\%.
PREN2D~\cite{yan2021primitive} slightly wins on CUTE, but IterNet shows huge performance gains on all the other datasets: 1.2\% on IIIT, 1.1\% on SVT, 1.5\% on IC13$_\text{S}$, 4.7\% on IC15$_\text{S}$, and 3.3\% on SVTP.

It's worth noting that our IterNet uses the same iterative language modeling module as ABINet, but with a different vision modeling module (\ie, IterVM). Compared with ABINet, our IterNet improved the accuracy by 0.6\%, 1.6\%, 0.5\%, 1.7\%, 1.6\%, and 1.1\% on IIIT, SVT, IC13, IC15, SVTP, and CUTE, respectively. On average, the accuracy improved from 95.6\% to 96.8\% for regular datasets and 87.2\% to 88.8\% for irregular datasets. 
These results indicate that the vision modeling module in the state-of-the-art scene text recognizer has become a performance bottleneck, and the language modeling module cannot address this problem alone, while our IterVM can effectively alleviate this problem.

\subsection{Qualitative Analysis}

\begin{table}[]
\centering
\caption{
Examples of the intermediate prediction results obtained by the proposed vision modeling module at different iterations. 
}
\vspace{-.0em}
\label{tab:visal_iterative}
\begin{tabular}{c|cc|c|c}
\toprule
\multirow{2}{*}{Image}         & \multicolumn{2}{c|}{Intermediate   Prediction}       & Final                    & Ground                   \\
                                                                                              & 1                        & 2                        & Prediction               & Truth                    \\
\midrule
\multirow{2}{*}{\includegraphics[width=0.25\linewidth]{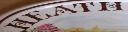}}         & \multirow{2}{*}{\textcolor{red}{sa}a\textcolor{red}{\_}h}    & \multirow{2}{*}{h\textcolor{red}{l}ath}   & \multirow{2}{*}{heath}   & \multirow{2}{*}{heath}   \\ &&&\\
\multirow{2}{*}{\includegraphics[width=0.25\linewidth]{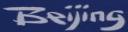}} & \multirow{2}{*}{b\textcolor{red}{oj}jing} & \multirow{2}{*}{\textcolor{red}{k}eijing} & \multirow{2}{*}{beijing} & \multirow{2}{*}{beijing} \\ &&&\\
\multirow{2}{*}{\includegraphics[width=0.25\linewidth]{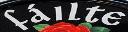}} & \multirow{2}{*}{\textcolor{red}{c}a\textcolor{red}{l}lt} & \multirow{2}{*}{fa\textcolor{red}{l}lte} & \multirow{2}{*}{failte} & \multirow{2}{*}{failte} \\ &&&\\
\multirow{2}{*}{\includegraphics[width=0.25\linewidth]{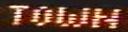}} & \multirow{2}{*}{to\textcolor{red}{l}n\textcolor{red}{h}} & \multirow{2}{*}{tow\textcolor{red}{h}} & \multirow{2}{*}{town} & \multirow{2}{*}{town} \\ &&&\\
\multirow{2}{*}{\includegraphics[width=0.25\linewidth]{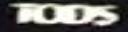}} & \multirow{2}{*}{\textcolor{red}{5520}} & \multirow{2}{*}{tod\textcolor{red}{5}} & \multirow{2}{*}{tods} & \multirow{2}{*}{tods} \\ &&&\\
\multirow{2}{*}{\includegraphics[width=0.25\linewidth]{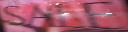}} & \multirow{2}{*}{saf} & \multirow{2}{*}{saee} & \multirow{2}{*}{sale} & \multirow{2}{*}{sale} \\ &&&\\
\multirow{2}{*}{\includegraphics[width=0.25\linewidth]{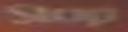}} & \multirow{2}{*}{s\textcolor{red}{\_u}p} & \multirow{2}{*}{s\textcolor{red}{w}op} & \multirow{2}{*}{stop} & \multirow{2}{*}{stop} \\ &&&\\
\multirow{2}{*}{\includegraphics[width=0.25\linewidth]{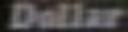}} & \multirow{2}{*}{do\textcolor{red}{i}l\textcolor{red}{z}r} & \multirow{2}{*}{do\textcolor{red}{i}lar} & \multirow{2}{*}{dollar} & \multirow{2}{*}{dollar} \\ &&&\\
 \bottomrule
\end{tabular}
\vspace{-.0em}
\end{table}

\subsubsection{Analysis on iterative correction by IterVM}
To have a comprehensive cognition about iterative correction by IterVM, we visualize the intermediate predictions on the challenging examples in Tab.~\ref{tab:visal_iterative}.
The predictions at the first iteration are always incorrect due to the low quality of the input image. After multiple iterations, IterVM can finally correct the predictions. 
It demonstrates the ability of our IterVM to refine text prediction iteratively.

\subsubsection{Result examples on challenging cases}
\begin{figure}[t]
    \centering
    \includegraphics[width=\linewidth]{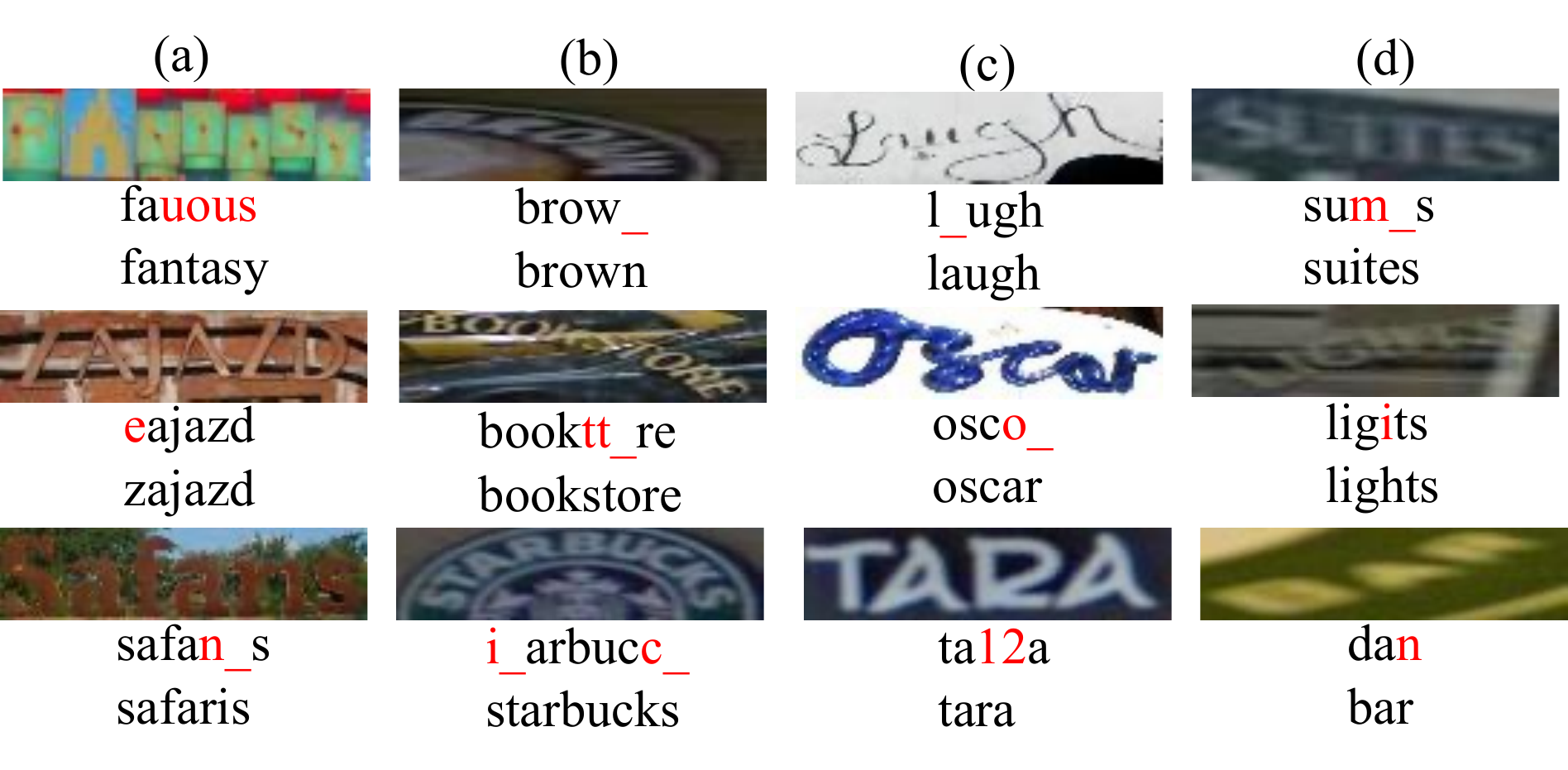}
    \caption{Challenging examples that ABINet fails (first line) but IterNet successes (second line): (a) visually ambiguous text, (b) heavily curved text, (c) special typefaces, and (d) low resolutions.}
    \label{fig:examples}
\vspace{-.0em}
\end{figure}

Fig.~\ref{fig:examples} depicts some challenging examples on which state-of-the-art method ABINet fails. As can be seen, IterNet successfully produces correct results on these challenging examples. 

\section{Conclusion}
This work explores the new idea of iterative vision modeling and proposes a novel vision modeling module named IterVM. Specifically, IterVM repeatedly uses the high-level visual feature extracted at the previous iteration to enhance the multi-level features extracted at the subsequent iteration. Based on IterVM, a powerful scene text recognizer called IterNet is further proposed, by combining it with iterative language modeling module (IterLM). Extensive comparisons and ablation studies are conducted, and the results show that: (1) the proposed IterVM is very effective for improving the scene text recognition accuracy, and (2) equipped with both IterVM and IterLM, the proposed IterNet is able to achieve new state-of-the-art results on public benchmarks.






\bibliographystyle{IEEEtran}
\bibliography{IEEEfull}
%



\end{document}